%% file: paper_template.tex
\documentclass[conference]{IEEEtran}
\IEEEoverridecommandlockouts
\usepackage{times}
\usepackage[numbers,compress]{natbib}
\usepackage{multicol}
\usepackage[bookmarks=true]{hyperref}
\usepackage{booktabs}
\usepackage{graphicx}
\usepackage{capt-of}
\usepackage{xcolor}
\usepackage{dblfloatfix}
\usepackage{booktabs,tabularx}
\usepackage{siunitx}
\usepackage{float}
\usepackage{amsmath}
\begin{document}

\title{UMI-Underwater: Learning Underwater Manipulation without Underwater Teleoperation}

\author{
Hao Li$^{*}$, Long Yin Chung$^{*}$, Jack Goler, Ryan Zhang, Xiaochi Xie, Huy Ha, Shuran Song, Mark Cutkosky
}

\makeatletter
\twocolumn[{%
\@twocolumnfalse
\maketitle

\begin{center}
  \includegraphics[width=0.9\textwidth]{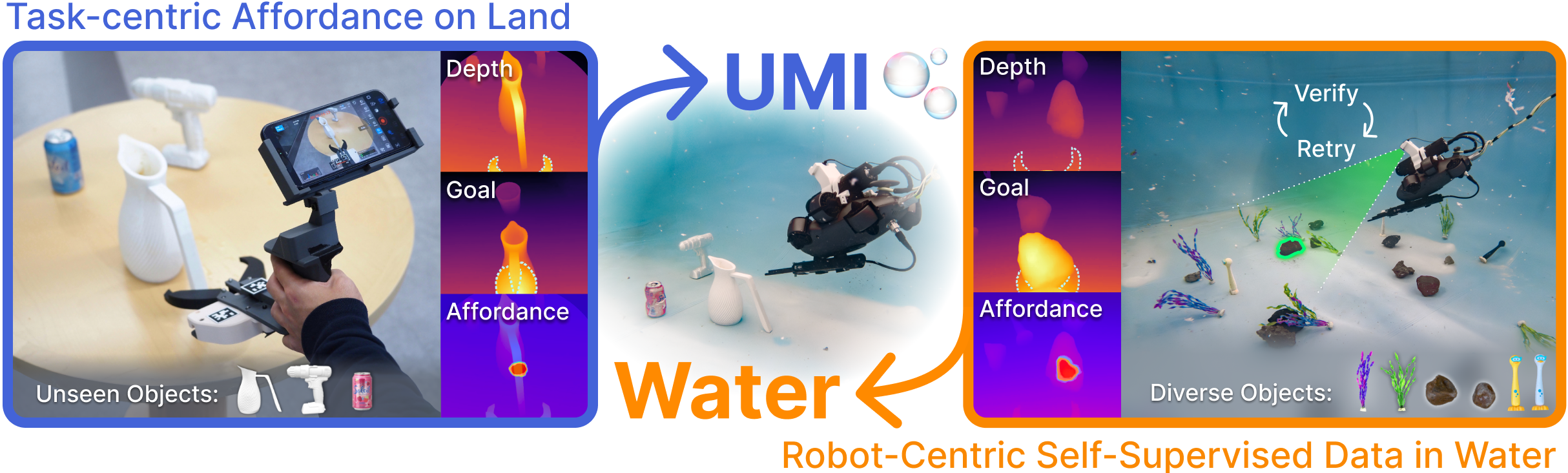}
  \captionof{figure}{\textbf{UMI-Underwater.} We address data-collection and generalization bottlenecks in underwater manipulation by pairing autonomous, self-supervised data collection with a zero-shot, depth-based affordance predictor that transfers directly from land to water.}
  \label{fig:overview}
\end{center}
\vspace{0.5em}
}]
\makeatother

\begingroup
\renewcommand\thefootnote{}
\footnotetext{%
$^{*}$Equal contribution. All authors are with Stanford University.
Emails: \{li2053, lychung, jgoler, rrzhang, xxie15, huyha, shuran, cutkosky\}@stanford.edu.
Correspondence to: li2053@stanford.edu.}
\endgroup

\begin{abstract}
\input{abstract}
\end{abstract}

\section{Introduction}
\label{sec:intro}

Underwater robotic manipulation is key to enabling applications such as ecological sampling, debris removal, and infrastructure inspection, yet robust autonomy remains challenging. Compared to terrestrial settings, underwater robots operate under severe perceptual degradation and variability---including wavelength-dependent attenuation, scattering/backscatter, turbidity, and rapidly changing illumination and caustics---that can substantially shift appearance even within the same environment \cite{akkaynak2018revised,akkaynak2019seathru,li2020uieb,anwar2019uieSurvey}. In addition, hydrodynamic disturbances and vehicle--object coupling complicate contact-rich interaction. These factors often make end-to-end visuomotor policies brittle under distribution shift, and they amplify the cost of collecting sufficient underwater training data.

A major practical bottleneck is the \emph{human burden in data collection}. Most underwater manipulation systems remain teleoperation-centric, but collecting diverse, high-quality underwater demonstrations is time-consuming and expensive. Recent work has begun to reduce this burden by leveraging self-supervised interaction to scale underwater data collection; by deploying autonomous heuristic controllers, systems can automatically acquire the successful demonstrations required to train robust visuomotor policies \cite{liu2024aquabot}. More broadly, self-supervised interaction has long been used to scale terrestrial robot data collection with automatically obtained success signals and repeated deployment \cite{pinto2015supersizing, levine2018handeye, zeng2018synergies, zeng2020tossingbot}.

In parallel, a growing line of work reduces demonstration burden by replacing robot-centric teleoperation with \emph{portable, robot-agnostic human demonstration interfaces}. Universal Manipulation Interface (UMI) enables in-the-wild data collection with a handheld gripper and transfers learned visuomotor policies to robots \cite{chi2024universal}. Follow-up systems extend UMI-style data collection and transfer to new embodiments and deployment constraints, including mobile manipulation on legged platforms, aerial manipulation under challenging dynamics, and more scalable/hardware-independent variants with larger datasets and richer sensing \cite{ha2024umi, gupta2025umi, zhaxizhuoma2025fastumi, liu2025fastumi, choi2026wild, rayyan2025mv, xu2025dexumi}.

This paper introduces a representation-centric method for robust underwater manipulation built around \emph{affordance}, while explicitly targeting human burden reduction in two complementary ways. \textbf{(1)} We introduce a self-supervised underwater data collection pipeline that autonomously collects successful grasp demonstrations to eliminate reliance on teleoperation. The collection system bootstraps grasp attempts with a heuristic controller, executes recovery behaviors to increase data efficiency, and filters episodes using an automatic success signal, yielding scalable underwater demonstrations. \textbf{(2)} We develop an UMI-derived handheld interface (\textbf{UMI-Aquatic}) to scalably collect diverse on-land demonstrations to train a depth-conditioned affordance model for grasp guidance. This allows us to (i) achieve \emph{zero-shot} land-to-water grasp knowledge transfer and (ii) bridge the land-to-water perception gap. Finally, we train an affordance and depth conditioned diffusion policy~\cite{chi2023diffusionpolicy} on action trajectories from our autonomous underwater collection system. This visuomotor policy captures the perception robustness of depth and the diverse grasp knowledge of affordance to improve generalization to novel manipulation targets and unseen environments.

We evaluate our approach on two regimes: (i) novel-object generalization and (ii) visual generalization to unseen backgrounds and lighting conditions. Across these settings, we find that depth-based affordances provide an effective perception interface for bridging land-to-water shift, and that autonomous data collection substantially reduces the human burden required to train underwater manipulation policies.

\noindent\textbf{Contributions:}
\begin{itemize}
    \item \textbf{Self-supervised underwater data collection:} a practical pipeline that autonomously gathers successful underwater grasp demonstrations using recovery behaviors and automatic success filtering.
    \item \textbf{Affordance heatmaps as a cross-domain perception interface:} a goal-conditioned, depth-based affordance predictor trained with on-land UMI-Aquatic demonstrations, enabling \emph{zero-shot} land-to-water transfer without mixed training or fine-tuning.
    \item \textbf{Robustness and transfer evaluation:} experiments on in-distribution pool grasping, background shifts, and novel-object generalization using objects seen only in the on-land dataset.
\end{itemize}

\section{Related Work}
\subsection{Underwater Manipulation and Learning}
\label{sec:rw_underwater}
Autonomous underwater manipulation has long been studied in the context of intervention missions (e.g., inspection, maintenance, and recovery), where the dominant challenges arise from hydrodynamic disturbances, limited visibility, and the complexity of controlling coupled vehicle--manipulator systems.
Early work and benchmark systems emphasized reliable mechatronic integration and robust control for free-floating intervention, including SAUVIM-style intervention platforms \citep{marani2009intervention}.
Subsequent efforts developed increasingly capable autonomy stacks for underwater vehicle--manipulator systems (UVMS), including the TRIDENT framework and its descendants \citep{sanz2010trident}, as well as large-scale projects such as DexROV that advanced semi-autonomous intervention and remote supervision \citep{dilillo2016dexrov}.
A complementary line of work focuses on principled control architectures, e.g., task-priority control for underwater intervention, to systematically handle competing objectives such as station keeping, collision avoidance, and end-effector motion \citep{simetti2018taskpriority}.

A complementary paradigm is human--robot haptic collaboration: Ocean One and the deep-sea OceanOneK use an ``avatar'' telepresence approach that enables dexterous bimanual manipulation under supervisory control, with haptic feedback as needed \citep{khatib2016ocean,brantner2021controlling,stuart2017oceanhands,oceanonek2022}.

Despite steady progress, many practical underwater manipulation systems remain teleoperation-centric, in part because collecting high-quality underwater demonstrations is expensive and time-consuming.
Learning from demonstration has therefore often been used to assist or partially automate underwater intervention, for example by learning task models that help disambiguate operator intent and improve resilience to communication constraints \citep{havoutis2019teleop}.
More recently, AquaBot showed that end-to-end visuomotor policies can be trained from demonstrations and improved through iterative self-learning beyond initial teleoperation performance \citep{liu2024aquabot}.

Building off this trajectory, our work addresses two complementary pillars of underwater manipulation: we replace expert teleoperation with an autonomous, self-supervised collection pipeline to scale data, while simultaneously exploring how transferable affordance representations can improve generalization across the land-to-water domain gap.

\subsection{Reducing human burden in demonstration collection}
\label{sec:related_burden}

A central challenge in learning-based manipulation is collecting diverse, high-quality demonstrations. In underwater settings this burden is amplified by limited visibility, vehicle--object coupling, and the operational overhead of teleoperation, motivating approaches that reduce (i) the amount of direct human-in-the-loop control and (ii) the effort required to collect diverse demonstrations.

\paragraph{Autonomous and self-supervised on-robot data collection}
A long line of work demonstrates that robots can scale interaction data by autonomously executing trials and using automatically obtained success signals for learning, substantially reducing manual labeling and teleoperation time\cite{pinto2015supersizing, levine2018handeye, zeng2018synergies, zeng2020tossingbot}.
Recent systems revisit this theme with practical autonomous data-collection pipelines that repeatedly deploy policies, accumulate experience, and update models under real-world constraints\cite{bousmalis2023robocat, papagiannis2024miles, mirchandani2024scaleup}.
Our underwater data collection pipeline follows this spirit: it autonomously generates grasp trials with recovery behaviors and retains successful episodes for behavior cloning, reducing reliance on underwater teleoperation.

\paragraph{Portable handheld interfaces and cross-embodiment transfer (UMI family)}
Complementary to autonomy-on-robot, handheld demonstration interfaces aim to make data collection cheap, portable, and robot-agnostic by capturing task-relevant trajectories with consistent observation/action semantics. Universal Manipulation Interface (UMI) enables in-the-wild demonstration collection using handheld grippers and transfers the learned visuomotor policies to robots without requiring robots at collection sites \cite{chi2024universal}.
Subsequent UMI-style systems extend this paradigm along multiple axes: transferring UMI policies to new embodiments with additional control constraints (e.g., legged mobile manipulation and aerial manipulation) \cite{ha2024umi, gupta2025umi}, redesigning the interface for easier deployment and larger-scale datasets \cite{zhaxizhuoma2025fastumi, liu2025fastumi}, augmenting the interface with contact-rich sensing (e.g., force/torque and tactile modalities) \cite{choi2026wild, cheng2026tacumi}, and broadening the interface concept to dexterous hands via wearable or vision-based adaptations \cite{xu2025dexumi}.
Our \textbf{UMI-Aquatic} interface is inspired by this line, but targets a distinct challenge: leveraging cheap on-land demonstrations to improve underwater manipulation, where teleoperation is especially time-consuming. We use a depth-based affordance representation and geometric alignment to enable zero-shot land-to-water transfer without mixed training or fine-tuning.

\subsection{Affordance Representations for Manipulation}
Affordances provide task-relevant, spatially localized cues for interaction. Early deep affordance approaches predict pixel-wise affordance labels or masks jointly with object detection \citep{do2018affordancenet}. Dense action-value or utility maps over image pixels have also been used to guide grasping and interaction \citep{zeng2018synergies}. Recent work studies affordances that generalize to novel objects, including few-shot affordance learning for unseen articulated objects \citep{ning2023where2explore}. In parallel, robust manipulation can benefit from strong intermediate visual representations such as monocular depth \citep{ranftl2021dpt,ranftl2022robustDepth}. Our contribution is to use an affordance heatmap as a modular perception interface that can be trained with on-land data, and then used to guide an underwater diffusion policy.

\section{Approach}
\label{sec:approach}

\subsection{Experimental Setup}
\label{sec:setup}

\begin{figure}[h]
    \centering
    \includegraphics[width=0.5\textwidth]{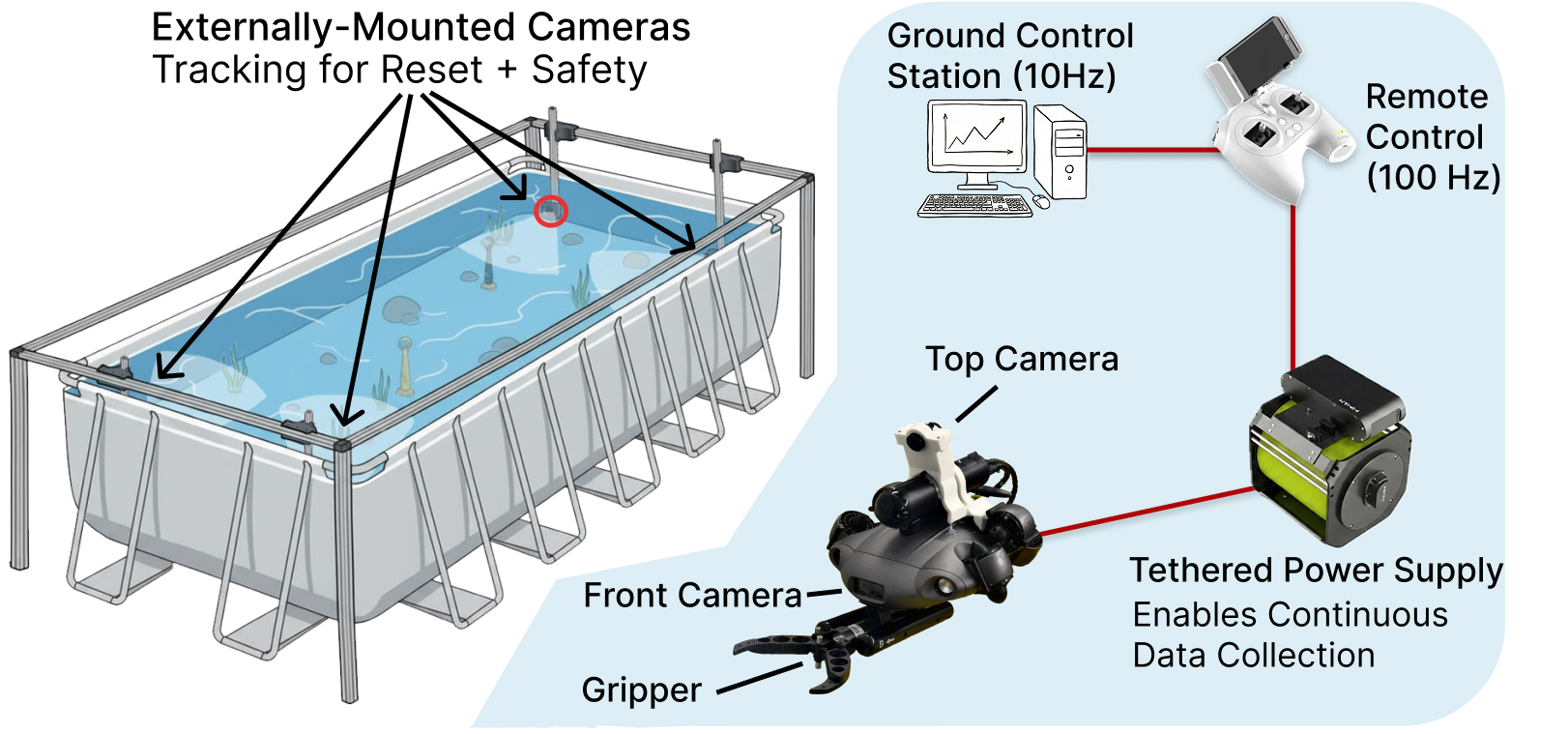}
    \vspace{-2em}
    \caption{\textbf{ROV Manipulation Setup for Self-supervised Underwater Data Collection.} The setup includes a swimming pool, with objects scattered across the workspace for repeated grasp attempts. Four fixed external cameras are mounted at the pool corners to provide real-time 3D localization used for safety functions (but not provided as inputs to the learned policy). The tethered ROV operates in this environment to execute the staged grasping routine.}
    \label{fig:system}
\end{figure}

Figure~\ref{fig:system} illustrates the overall system. The ROV manipulation setup is similar to that in prior autonomous underwater manipulation (e.g., \citep{liu2024aquabot}), with updates enabled by recent improvements to the QYSEA platform SDK.

\paragraph{Platform}
The platform is a compact tethered underwater ROV equipped with a forward-facing RGB camera, an additional wide-FOV RGB camera mounted at the top, and a 1-DoF gripper (open/close). The vehicle is operated through a tethered control box at 100\,Hz. For autonomous operation, the vendor SDK is used to send velocity and gripper commands and to receive onboard proprioception (IMU and depth).

\paragraph{Cameras}
Earlier reports noted high latency from the integrated front camera and therefore added an external low-latency camera \citep{liu2024aquabot}. With the newer SDK, front-camera latency is substantially reduced (below 100\,ms). An external waterproof camera is nevertheless retained for two practical reasons: (i) it provides a complementary viewpoint to the onboard cameras, and (ii) the wide-FOV view typically offers richer scene context during approach and grasp. The external camera is mounted via a 3D-printed bracket rigidly attached to the vehicle body.

\paragraph{State estimation and safety}
The environment is equipped with four fixed external cameras placed at the pool corners for real-time 3D localization of the vehicle. Fusing the externally estimated 3D position with onboard IMU yields a 6-DoF pose estimate in a global frame. This position estimate is \emph{not} provided to the learned policy; it is reserved for safety (collision avoidance), episode reset, and operator monitoring.

\paragraph{Control constraints}
During both data collection and policy execution, vehicle pitch is fixed to a constant downward angle (10$^\circ$) to stabilize the camera viewpoint and simplify the grasping geometry. This constraint reduces control complexity and improves repeatability while remaining representative of close-range grasping behaviors in this setting.

\begin{figure}[h]
    \centering
    \includegraphics[width=0.5\textwidth]{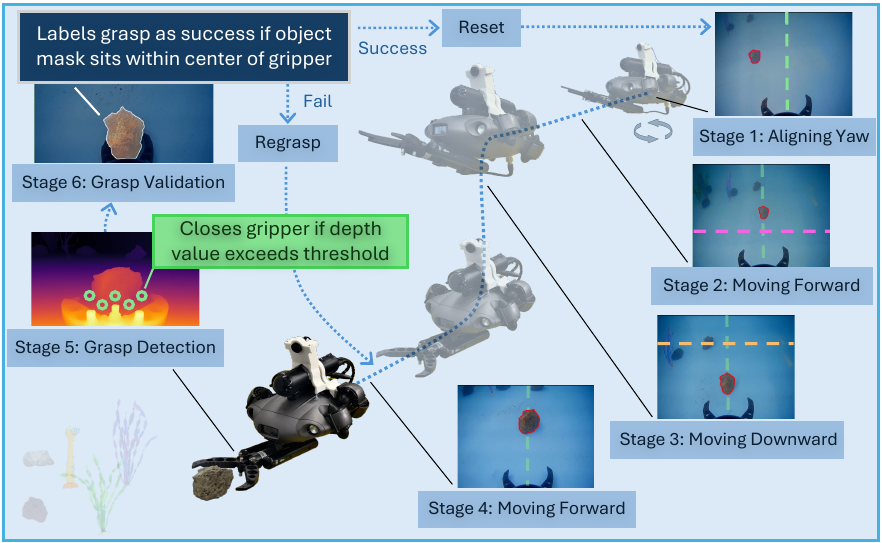}
    \vspace{-2em}
    \caption{\textbf{Autonomous Data Collection Pipeline.} Our heuristic controller autonomously collects grasping episodes by using a segmentation model to select a target, servoing the object centroid to stage-specific pixel setpoints using PD control, closing the gripper when a depth threshold is met, and labeling success via drag validation.}
    \label{fig:heuristic_control}
\end{figure}
\vspace{-1.5em}

\subsection{Data Collection}
\label{sec:selfsup}
A key challenge in underwater learning is acquiring large quantities of robot interaction data without relying on extensive teleoperation. We therefore design a self-supervised data collection pipeline that uses a heuristic visuo-servoing PD controller to autonomously generate grasp trajectories. Each episode proceeds through a sequence of stages (Figure~\ref{fig:heuristic_control}).

\paragraph{Reset and initialization}
At the start of each episode, the vehicle is placed at a shallow depth with a random initial pose that provides a wide field of view over the workspace. Using the external tracking system, we initialize the vehicle state and enforce the fixed pitch constraint.

\paragraph{Object selection and tracking}
We run real-time segmentation with (SAM2-tiny \citep{ravi2024sam}) to obtain per-frame object masks. From the visible objects, we select the target whose centroid is closest to the image center and track its mask across frames. We compute a target pixel as the centroid of the selected mask.

\paragraph{Stage 1: Yaw alignment}
We yaw the vehicle such that the image centerline coincides with the target centroid. Specifically, we define the horizontal pixel error between the centroid and the image centerline and use a PD controller to command yaw until the error falls below a threshold. It remains active throughout the grasp sequence to maintain target centering during approach and closure.

\paragraph{Stage 2: Forward approach}
After initial yaw alignment, we command forward motion to reduce the apparent distance to the object. Concretely, we define a reference horizontal line in the lower portion of the image and continue moving forward until the centroid reaches this reference line (i.e., until the vertical pixel error meets a threshold). This provides a simple but effective proxy for range without requiring a calibrated stereo system.

\paragraph{Stage 3: Depth adjustment}
We then adjust depth to keep the target within a favorable grasping region. We define a second reference horizontal line in the upper portion of the image and command vertical motion until the centroid lies within the target band.

\paragraph{Stages 4 \& 5: Close-range approach and grasp}
Once the vehicle is sufficiently close, we use a monocular depth estimator (Depth Anything V2 \citep{yang2024depth}) on the onboard RGB stream to further regulate approach distance. When the estimated distance indicates that the object is within the gripper workspace, the gripper closes to attempt a grasp.

\paragraph{Stage 6: Drag verification}
To verify grasp success without external instrumentation, we execute a short retreat motion (dragging the object) for 3\,s. If the object remains in the gripper without slipping during this interval, we label the episode as a success; otherwise as a failure.

\begin{figure}[h]
    \centering
    \includegraphics[width=\columnwidth]{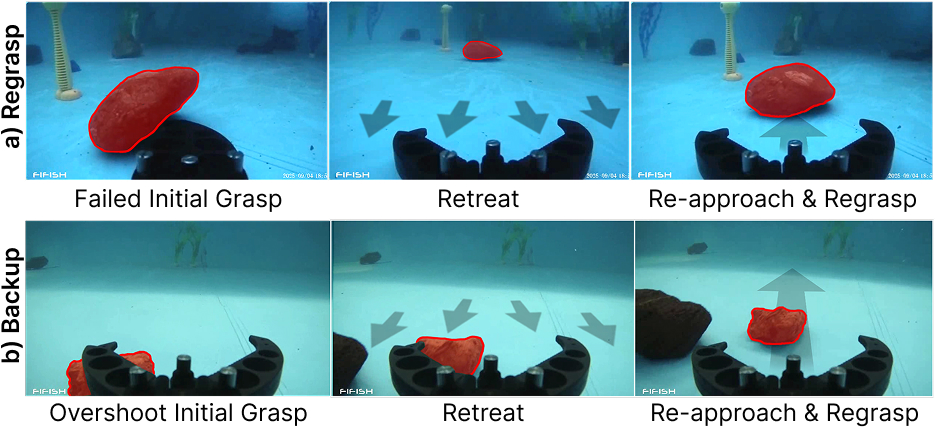}
    \vspace{-2em}
    \caption{\textbf{Autonomous Recovery Strategies} include \textbf{(a) regrasp} after failed grasps and \textbf{(b) backup} when the robot overshoots. These strategies improve the success rate of demonstration collection as well as improve policy robustness by demonstrating recovery behavior.}
    \label{fig:recovery}
\end{figure}

\paragraph{Failure recovery mechanisms}
We incorporate two recovery behaviors that substantially improve autonomous collection (Fig. \ref{fig:recovery}) :
(1) \textbf{Regrasp.} After each failed grasp attempt, the vehicle backs up, applies a small lateral offset (randomly left/right), reopens the gripper and retries the approach. This increases the probability of completing a successful grasp after poor initial contact.
(2) \textbf{Overshoot.} Due to hydrodynamic effects, the vehicle can overshoot, moving the object toward the image boundary or out of view. We detect imminent loss-of-view when the target centroid approaches the image margin; in that case we command a larger retreat to re-acquire the object before restarting alignment.

\begin{figure}[h!]
    \centering
    \includegraphics[width=0.9\linewidth]{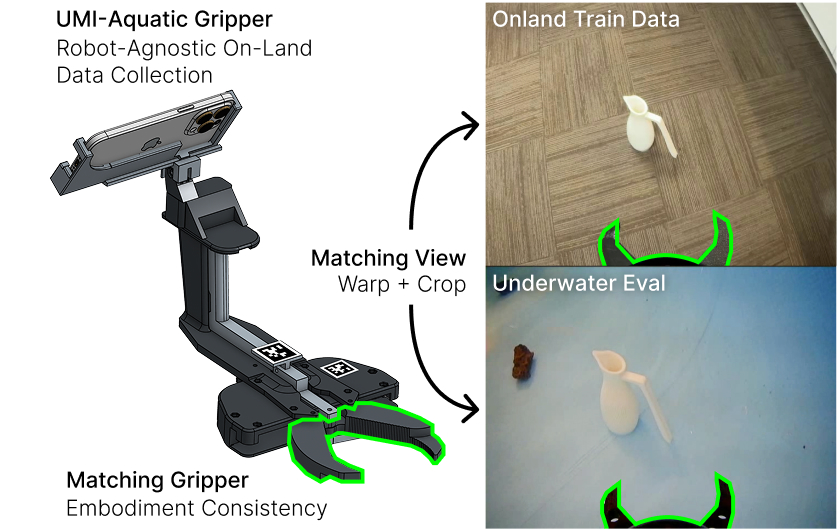}
    \vspace{-0.8em}
    \caption{\textbf{UMI-Aquatic on-land demonstration setup.} Our handheld gripper with an iPhone camera system and AprilTags enables portable data collection and reliable gripper-state tracking for automatic demonstration labeling. Cropping and geometric warping via reprojection align the iPhone view to match the underwater robot camera.}
    \label{fig:gripper}
\end{figure}

\paragraph{On-land demonstrations with UMI-Aquatic}
To collect diverse grasping demonstrations on land, we introduce \textbf{UMI-Aquatic}, a handheld gripper interface derived from the Universal Manipulation Interface (UMI)~\citep{chi2024universal}. UMI-Aquatic integrates (i) an iPhone camera rigidly mounted to match the underwater camera viewpoint and (ii) AprilTags mounted on the gripper to enable reliable gripper-state tracking during handheld operation (Fig.~\ref{fig:gripper}). Compared to the original UMI design, our modifications are tailored to underwater transfer: the camera mounting is designed to simplify cross-camera alignment, and the tag-based tracking provides a robust estimate of gripper openness/closure timing used for automatic label generation.

\paragraph{Scale and human effort}
Our self-supervised pool pipeline collected 536 grasp demonstrations, with each episode lasting $\sim$45\,s and between-episode reset taking $\sim$60\,s, for a total of $\sim$15\,h of autonomous runtime for the episodes plus resets. From this set, we used 233 successful underwater grasps to train the diffusion policy.

Objects were re-scattered roughly every 5\,h, and manual intervention was only required when the vehicle became stuck or the tether tangled.
Separately, UMI-Aquatic enabled fast on-land collection of 800 handheld demonstrations across 6 object types and 8 backgrounds (each $\sim$10\,s, $\sim$2.2\,h of active demonstrating)

\subsection{Goal-Conditioned Data Preprocessing and Training}
\label{sec:dp_affordance}

We improve robustness and generalization by separating \emph{where to grasp} from \emph{how to control}. We learn (i) a goal-conditioned affordance predictor that outputs a dense grasp heatmap and (ii) a visuomotor policy that conditions on this heatmap (and depth) to produce control commands. To reduce sensitivity to lighting and color differences between land and underwater scenes, our affordance predictor operates on depth rather than RGB; in our setting, RGB-input affordance models trained on on-land UMI-Aquatic data transfer poorly to underwater imagery due to severe appearance shift.

\paragraph{Training tuples and goal-conditioning}
From each episode we construct training tuples $\{(D_t, D_{\mathrm{goal}}, \hat{H}_t)\}_{t=1}^{T}$, where $D_t$ is the depth observation at time $t$, $D_{\mathrm{goal}}$ is a goal depth image, and $\hat{H}_t$ is a sparse per-pixel affordance target. Goal-conditioning disambiguates which object to grasp in multi-object scenes.

\emph{Goal source.} In our setting, the goal observation is provided \textbf{from on-land UMI-Aquatic data}: at test time we specify the target with a single on-land goal RGB image, convert it to a depth map using the same monocular depth pipeline, and use it as $D_{\mathrm{goal}}$ (after the same crop/warp preprocessing as in Sec.~C.~d). Thus, the affordance model performs \textbf{cross-domain goal matching} between the current underwater observation and an on-land goal, and we do \textbf{not} require capturing an underwater goal image at deployment.

\begin{figure}[h]
    \centering
    \includegraphics[width=0.9\linewidth]{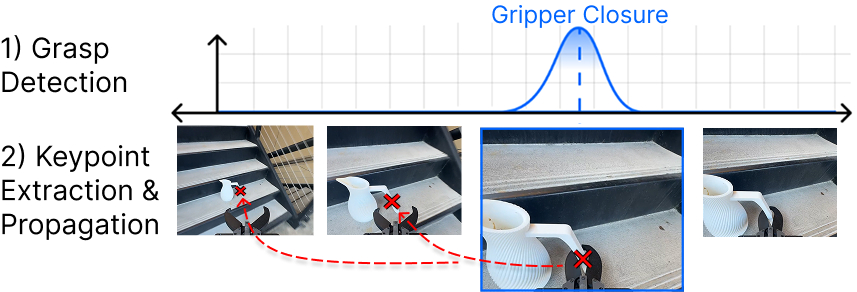}
    \vspace{-0.5em}
    \caption{\textbf{Automatic Affordance Supervision.} AprilTag tracking identifies gripper closure timing using a sliding window, which seeds point tracking to backtrack grasp contact keypoints on the object for affordance supervision.}
    \label{fig:affordance_labeling}
\end{figure}

\paragraph{Automatic per-frame supervision}
For on-land data collected with UMI-Aquatic as shown in Fig. \ref{fig:gripper}, we use an AprilTag-based gripper tracking pipeline (Fig. \ref{fig:affordance_labeling}) to estimate gripper openness/width and to determine the grasp/contact timing in the camera frame. We detect the closure time $t^\star$ using a sliding-window change detector over the width signal (a sharp drop followed by a stable plateau). Then, we backtrack the gripper--object contact pixel from the first frame where the gripper is closed, $t^\star$, to earlier frames using point tracking (CoTracker3 \cite{karaev2025cotracker3}), yielding a tracked contact location $(u_t, v_t)$ for each $t \le t^\star$, producing compatible targets $\hat{H}_t$.

\paragraph{Land-to-water pretraining via diverse on-land data (zero-shot transfer)}
To improve generalization, we train an affordance model on a diverse on-land dataset. An iPhone-based capture setup is portable and easy to deploy, enabling rapid collection of varied interactions without specialized site preparation (e.g., no mapping run). Because the iPhone camera differs from the underwater camera in intrinsics, distortion, and field-of-view, we reduce this geometric domain gap by calibrating both camera models and warping on-land frames into the underwater camera geometry. Using depth as the affordance input further reduces the perception gap caused by lighting and color differences. Importantly, there is \emph{no mixed training and no fine-tuning}: online affordance deployment uses on-land data only while offline affordance used for training the policy uses underwater data only.

\begin{figure*}[ht!]
    \centering \includegraphics[width=0.9\textwidth]{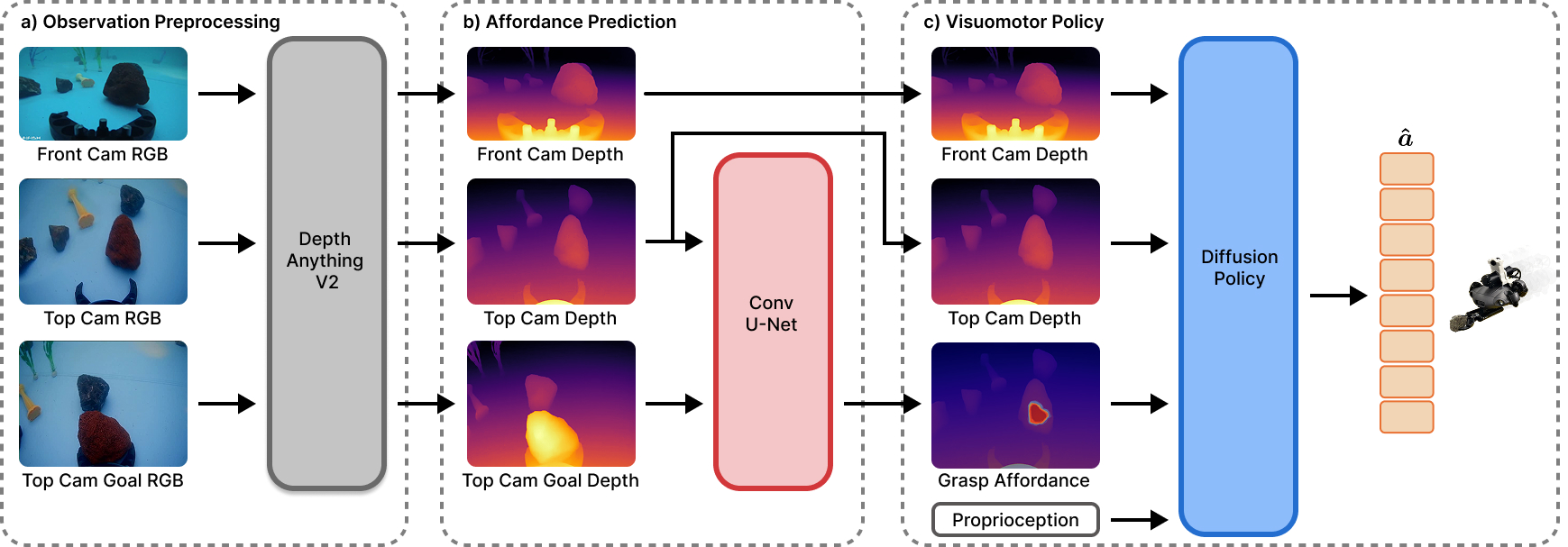}
    \vspace{-0.5em}
    \caption{\textbf{Model Architecture.} \textbf{(a) Observation preprocessing} takes monocular RGB current and goal observations from robot cameras and converts them to depth maps using Depth Anything V2. \textbf{(b) Affordance prediction} takes top camera depth and goal depth observations and uses a convolution U-Net to predict a dense grasp affordance heatmap for target localization. The online affordance model for underwater deployment is trained on exclusively on-land UMI-Aquatic data and is deployed underwater zero-shot. We supervise affordance training using automatically labeled keypoint maps that indicate the ground truth grasp points. \textbf{(c) Visuomotor policy} conditions on the predicted affordance map, depth observations, and robot proprioception to generate a short-horizon action trajectory.}
    \label{fig:model}
\vspace{-2em}
\end{figure*}

\paragraph{Affordance dataset preparation and cross-camera warping}
\label{sec:warp}
A practical challenge for land-to-water transfer is the mismatch between the iPhone camera used for on-land collection and the waterproof wide-view camera used underwater. We therefore pre-warp each on-land frame into the underwater camera image geometry using a calibrated, plane-at-depth remapping.

\smallskip
\noindent\textbf{Plane-at-depth warp (iPhone $\rightarrow$ underwater camera).}
Let camera~1 denote the iPhone (source) and camera~2 denote the underwater camera (target), with pinhole intrinsics $K_1, K_2$ and distortion parameters $\mathbf{d}_1, \mathbf{d}_2$. For each target pixel $\tilde{\mathbf{p}}_2=(u_2,v_2)$ in camera~2, we undistort to normalized coordinates $(x_2,y_2)$ and form a ray $\mathbf{r}_2=[x_2,y_2,1]^\top$. We intersect this ray with a fronto-parallel plane at depth $Z$ in camera~2 coordinates:
\begin{equation}
\mathbf{X}_2 = Z\,\mathbf{r}_2.
\end{equation}
We then transform to camera~1 coordinates and project into the iPhone image:
\begin{align}
\mathbf{X}_1 &= R\,\mathbf{X}_2 + t, \\
\tilde{\mathbf{p}}_1 &= \pi\!\left(K_1, \mathbf{d}_1, \mathbf{X}_1\right),
\end{align}
where $\pi(\cdot)$ denotes perspective projection with lens distortion. Since we mount the iPhone camera to match the underwater camera orientation, we use $R=I$ (translation-only alignment). Finally, we perform backward sampling:
\begin{equation}
I_{1\rightarrow2}(\tilde{\mathbf{p}}_2) = I_1(\tilde{\mathbf{p}}_1),
\end{equation}
implemented as a dense remapping table applied with bilinear resampling for images.

\paragraph{Affordance model architecture and training}
We train a goal-conditioned affordance predictor from depth sequences. For each training sample, we take the current depth image $D_t$ and a goal depth image $D_{\mathrm{goal}}$, normalize depth to $[0,1]$ using dataset-wide anchored min/max statistics (with clamping), resize to $112{\times}112$, and concatenate $(D_t, D_{\mathrm{goal}})$ into a two-channel input. Targets are per-pixel keypoint maps representing ground truth grasp locations, resized with nearest-neighbor interpolation. Our model is a U-Net (depth 4, base channels 64) with convolution--BN--ReLU encoder blocks and max-pooling, a bottleneck with increased channel capacity, and a mirrored decoder with transposed-convolution upsampling and skip connections. A final $1{\times}1$ convolution outputs per-pixel logits at the input resolution. Training uses a weighted BCE-with-logits objective to address foreground/background imbalance, AdamW optimization with warmup+cosine scheduling, gradient clipping, and mixed precision, with depth-specific augmentations (geometric transforms plus scale/noise/dropout/blur). We use episode-wise train/validation splits to avoid temporal leakage.

\smallskip

\subsection{Affordance Conditioned Visuomotor Policy}

\paragraph{Policy inputs}
The diffusion policy receives a multi-modal input with: (i) predicted affordance heatmap, (ii) monocular depth map predicted from RGB, and (iii) low-dimensional robot status signals (compass, pitch, vehicle depth). During policy training, we generate affordance heatmaps by training an affordance predictor on the same self-supervised underwater training dataset and applying it offline to all demo frames. We do not use external motion-capture pose for policy input; it is reserved for safety and reset.

\paragraph{Diffusion policy}
We adopt a diffusion policy that models a horizon of future actions and generates commands via iterative denoising. Observations are encoded by a pretrained vision backbone (ViT-B/16 CLIP \cite{radford2021learning}) into a global latent condition, and a 1D UNet denoiser predicts a sequence of actions. We predict a 16-step action horizon with a 6-dimensional action vector corresponding to yaw, forward/backward, up/down, left/right, open, close, executed in a receding-horizon manner.

\paragraph{Training and compute}
We train the diffusion policy by behavior cloning on successful demonstrations using the standard diffusion objective (MSE on predicted noise for noised action trajectories). Training is performed on a single NVIDIA RTX 5090 GPU with batch size 56 for 150 epochs.

\paragraph{Inference and deployment}
At test time, we run the affordance predictor, monocular depth estimation, and diffusion policy on a desktop equipped with an NVIDIA RTX 5090 GPU and an AMD Ryzen 9800X3D CPU. The full perception--control stack runs online and publishes actions at 10 Hz, with policy inference updates at 1 Hz.

\section{Experiments}
\label{sec:experiments}

\subsection{Evaluation Setup}
\label{sec:eval_setup}

\paragraph{Environment}
All experiments are conducted in an above-ground swimming pool, approximately 2$\times$4\,m in area and 1\,m deep. Unless otherwise stated, objects are placed within a fixed workspace region on the pool floor and the robot executes a pick-and-drag grasping episode of fixed duration.

\paragraph{Task}
Our primary task is \emph{pick-and-drag grasping}: the robot must grasp a target object and maintain grasp while retreating for a fixed duration.

\paragraph{Multi-object evaluation and goal specification}
In each trial, we place three objects in front of the robot (for both ID and OOD settings). All three objects are visible in the robot's initial camera view. We specify the desired target using a single on-land goal image from UMI-Aquatic: the operator selects an on-land goal RGB image of the intended target object, which is then converted to a goal depth map $D_{\mathrm{goal}}$ using the same monocular depth pipeline and preprocessing described in Sec.~\ref{sec:dp_affordance}. This goal is consumed by the \emph{affordance model} to produce a goal-aligned heatmap used by affordance-conditioned policies. RGB-only baselines do not receive the goal image.

The main challenge is therefore not only executing a grasp, but also selecting the correct target among distractors under appearance variation and cross-domain goal specification (on-land $\rightarrow$ underwater).

\paragraph{Object sets and evaluation settings}
We define a seen underwater object set used to collect pool demonstrations and train policies (rocks, seagrass, toy ducks), and a novel on-land UMI object set used only for evaluating novel-object transfer (pitcher with handle, can, power drill). We evaluate three settings:
\begin{enumerate}
    \item \textbf{In-distribution (ID):} seen underwater objects under the \emph{training} pool background distribution.
    \item \textbf{OOD visual generalization (background shift):} seen underwater objects under \emph{unseen} pool background wallpapers.
    \item \textbf{OOD novel-object generalization (UMI objects):} on-land UMI objects evaluated in the pool under the \emph{training} pool background distribution. This evaluation is zero-shot. There is no policy fine-tuning at this evaluation.
\end{enumerate}

\paragraph{Metric}
We report success rate (\%, higher is better): an episode is successful if the robot grasps the specified target object and it does not slip during a 3\,s retreat/drag verification (Section~\ref{sec:selfsup}). Grasping a non-target object is counted as a failure. For each method and condition, we run 20 trials.

\subsection{Methods Compared}
We compare diffusion-policy controllers trained on successful pool demonstrations, with different perceptual inputs and affordance pretraining sources. Let \textbf{RGB} denote raw RGB observations, \textbf{Depth} denote monocular depth predicted from RGB, and \textbf{Aff} denote the learned goal-conditioned affordance heatmap. All diffusion policies are trained on the same pool demonstration set (seen underwater objects under the training pool background), and differ only in their observation modalities and/or the affordance model used.

\paragraph{Goal inputs and fairness}
For methods that use \textbf{Aff}, the goal image is provided \emph{only} to the affordance model to produce the heatmap; the diffusion policy itself is conditioned on the predicted heatmap (plus depth and proprioception), not directly on the goal image. The \textbf{DP+RGB} baseline is \textbf{goal-agnostic} and represents the common setting of training an end-to-end underwater policy from in-water demonstrations without cross-domain goal-conditioning. Since \textbf{DP+RGB} receives no explicit goal specification in multi-object scenes, we define its implicit target as the object closest to the image center in the initial view, matching the center-biased target selection used during our self-supervised data collection pipeline (Section~\ref{sec:selfsup}). To isolate the effect of adding RGB while keeping goal-conditioning fixed, we compare \textbf{DP+Aff+Depth} against \textbf{DP+RGB+Aff+Depth}, which both use the \emph{same} (goal-conditioned) affordance model.

\paragraph{Important note on affordance supervision}
Although it is possible to train an affordance model using only pool-collected self-supervised data, in our setting this pool-only dataset is highly biased: our underwater data collection follows a multi-stage heuristic controller that repeatedly approaches objects with similar trajectories and viewpoints, and the dataset is relatively small and dominated by a narrow subset of objects (e.g., rocks). As a result, pool-only affordance training produces less robust affordance predictions. Our method uses UMI-Aquatic to collect on-land demonstrations of the pool object categories (rocks, seagrass, toy duck), trains the affordance model on this dataset, and then deploys it underwater in a zero-shot manner.

\paragraph{Compared methods}
\begin{itemize}
    \item \textbf{DP + Aff + Depth (ours) (UMI-Aquatic-pretrained Aff)}: diffusion policy conditioned on the affordance heatmap and monocular depth (and proprioception), trained on pool demonstrations from the seen-object set; the affordance model is trained on on-land UMI-Aquatic demonstrations and transferred underwater zero-shot (no mixed training, no fine-tuning).

    \item \textbf{DP + RGB (goal-agnostic; seen-only)}: diffusion policy conditioned on RGB (and proprioception), trained on pool demonstrations from the seen-object set (rocks/seagrass/duck).

    \item \textbf{DP + RGB + Aff + Depth (RGB-addition ablation; UMI-Aquatic-pretrained Aff)}: diffusion policy conditioned on RGB, affordance heatmap, and depth (and proprioception), trained on pool demonstrations; the affordance model is the same UMI-Aquatic-pretrained model as above.
\end{itemize}

\paragraph{RGB-addition ablation}
We include \textbf{DP + RGB + Aff + Depth} as an ablation to test whether providing \emph{raw RGB} in addition to the robust (Aff+Depth) interface helps or hurts. Because the brittleness of RGB is most pronounced under large appearance shift, we primarily report this ablation under the unseen-background setting (Table~\ref{tab:ood_bg}) to isolate the effect of adding RGB while keeping affordance and depth inputs fixed.

\paragraph{Train/test protocol}
All diffusion policies are trained on pool demonstrations collected with the seen underwater object set under the training pool background distribution. We evaluate: (i) in-distribution (ID) performance on seen objects under the training background, (ii) OOD visual generalization on seen objects under unseen background wallpapers, and (iii) OOD novel-object generalization on novel objects placed in the pool (zero-shot; no policy fine-tuning at test time).

\subsection{Results on the In-Distribution Task and Environment}
\label{sec:results_id}
We first evaluate in-distribution performance on the seen underwater object set under the training pool background, establishing a baseline before any distribution shift.

\begin{table}[t]
  \centering
  \caption{ID performance (seen objects + training pool background).}
  \label{tab:id}
  \begin{tabular}{lc}
    \toprule
    \textbf{Method} & \textbf{Success (\%)} $\uparrow$ \\
    \midrule
    DP + Aff + Depth (ours) & \textit{85} \\
    DP + RGB & \textit{65} \\
    \bottomrule
  \end{tabular}
\end{table}

As shown in Table~\ref{tab:id}, in the ID setting, \textbf{DP + Aff + Depth} outperforms the RGB-only baseline. The primary failure mode of \textbf{DP + RGB} in multi-object scenes is target confusion: it sometimes approaches and grasps a distractor object instead of the intended one. In contrast, the goal-conditioned affordance heatmap provides an explicit spatial goal signal that consistently focuses the policy on the desired target.

\subsection{OOD Visual Generalization Under Unseen Backgrounds}
\label{sec:results_visual_gen}
To evaluate visual robustness, we change pool background wallpapers while keeping the task and object set fixed (seen objects).

\begin{table}[b]
  \centering
  \caption{OOD generalization (seen objects + unseen backgrounds).}
  \label{tab:ood_bg}
  \begin{tabular}{lc}
    \toprule
    \textbf{Method} & \textbf{Success (\%)} $\uparrow$ \\
    \midrule
    DP + Aff + Depth (ours) & \textit{80} \\
    DP + RGB & \textit{0} \\
    DP + RGB + Aff + Depth & \textit{0} \\
    \bottomrule
  \end{tabular}
\end{table}

Table~\ref{tab:ood_bg} shows that under background shift, RGB-based policies fail catastrophically. The training pool background is visually simple and dominated by blue tones, whereas the wallpaper backgrounds introduce strong texture and color statistics (e.g., wood-grain patterns) that lie outside the training distribution (see supp.~material). As a result, policies that only take RGB observations achieve 0\% success. Notably, \textbf{DP + RGB + Aff + Depth} uses the \emph{same} affordance model and the \emph{same} depth input as \textbf{DP + Aff + Depth}; the only difference is that it also receives RGB. Its collapse to 0\% indicates that adding RGB can cripple robustness under strong appearance shift even when other robust signals are available. In contrast, \textbf{DP + Aff + Depth} remains robust: it relies on a goal-aligned affordance signal and depth cues that are less sensitive to background appearance, enabling reliable target selection and grasping under large visual shifts.

\subsection{Novel-Object Generalization via On-Land Pretraining}
\label{sec:results_novel_obj}
We evaluate novel-object generalization by testing in the pool on objects that appear only in the on-land UMI dataset (pitcher with handle, can, power drill), which are disjoint from the underwater seen-object set used to train diffusion policies (rocks/seagrass/duck). Our method uses an affordance model trained on on-land UMI grasping data and transfers it underwater zero-shot, while the DP+RGB baseline is trained only on seen underwater objects.

\begin{table}[t]
  \centering
  \caption{OOD novel-object generalization (Unseen objects in pool + training pool background), zero-shot.}
  \label{tab:novel-obj}
  \begin{tabular}{lc}
    \toprule
    \textbf{Method} & \textbf{Success (\%)} $\uparrow$ \\
    \midrule
    DP + Aff + Depth (ours) & \textit{75} \\
    DP + RGB & \textit{50} \\
    \bottomrule
  \end{tabular}
\vspace{-2em}
\end{table}

As reported in Table~\ref{tab:novel-obj}, on novel objects, \textbf{DP + Aff + Depth} substantially outperforms \textbf{DP + RGB}. Interestingly, the RGB-only policy still achieves non-trivial success (50\%), which we attribute to two factors: (i) the background remains unchanged from training, so many low-level visual statistics match the training distribution; and (ii) the novel objects share some coarse geometric properties with the seen set, allowing occasional grasps despite lacking explicit goal-aligned localization. However, the affordance-conditioned policy is markedly more robust, suggesting that on-land UMI-pretrained affordances provide a stronger inductive bias for selecting graspable regions on previously unseen objects.

\subsection{Failure mode: overshoot-induced target switch.}
While the affordance heatmap substantially reduces target confusion in multi-object scenes, we observed a recurring failure mode when the robot \emph{overshoots} during an approach. In these cases, the camera viewpoint can shift such that the target object moves partially out of view or becomes severely misaligned with the original goal specification. The affordance model may then place high confidence on a different salient region that remains visible (often a distractor object), which in turn guides the policy toward the wrong object. This failure can also \emph{invalidate our recovery strategy}: because recovery assumes the affordance remains anchored to the intended target, an overshoot-induced target switch can cause recovery to re-approach the distractor rather than returning to the original goal. We expect this issue can be mitigated by improved low-level control (reducing overshoot), and/or by adding temporal consistency to target localization (e.g., goal tracking or history-conditioned affordance prediction).

\section{Conclusion}
\label{sec:conclusion}
We presented a practical pipeline for underwater manipulation that combines (i) self-supervised underwater data collection for policy training and (ii) a goal-conditioned affordance model enabled by UMI-Aquatic to improve robustness and target selection. Our key idea is to separate \emph{goal localization} from \emph{control}: the affordance model provides a goal-aligned spatial heatmap, and a policy conditioned on affordance and depth executes pick-and-drag grasping. In experiments, this design improves in-distribution performance by reducing target confusion in multi-object scenes, and it increases robustness under large appearance shifts such as unseen pool wallpapers. Moreover, by transferring an affordance model trained on on-land data, our method generalizes to novel objects underwater without policy fine-tuning, demonstrating that on-land pretraining can effectively support underwater deployment when paired with appropriate conditioning.

\paragraph{Limitations and future work.}
A current limitation is that our affordance predictor is conditioned primarily on \emph{predicted depth} (from Depth Anything V2), which helps mitigate the domain gap between on-land and underwater RGB appearance, but can discard informative cues available in color and texture (e.g., specularities, material properties, object boundaries under poor depth estimates, and subtle target identity cues). This trade-off may be suboptimal in scenes where depth is noisy, where objects have similar geometry but different visual attributes, or where grasp success depends on material-specific cues.
A promising direction is to incorporate \emph{appearance information} while preserving domain robustness, e.g., via depth-first fusion with RGB features that are explicitly regularized for invariance (domain adaptation, style randomization, or contrastive objectives), or by using underwater-specific depth/geometry estimation and uncertainty-aware conditioning. More broadly, expanding the diversity of underwater demonstrations and object categories, and extending to longer-horizon tasks beyond pick-and-drag, could further improve generalization and unlock more capable underwater manipulation systems. Another limitation arises from low-level control: our platform currently uses a simple PID controller for motion execution, which can exhibit overshoot during approach and retreat motions. This overshoot not only reduces grasp reliability but can also induce the viewpoint changes that trigger the target-switch failure mode described in Section~\ref{sec:experiments}. A promising direction is to replace PID with a dynamics-aware controller such as MPC that explicitly accounts for system dynamics, actuation limits, and underwater drag, which could improve trajectory tracking and reduce overshoot.

\section*{Acknowledgments}
This work was supported in part by the NSF Award \#2143601, \#2037101, and \#2132519, and Samsung. We thank Tian-Ao Ren, Juhyun Jung, Stanley Wang, and Tianyu Tu for their thoughtful discussions and help with the experiments. We also give special thanks to Stanford Hopkins Marine Station for supporting our in-the-wild ocean deployment tests. The views and conclusions contained herein are those of the authors and should not be interpreted as necessarily representing the official policies, either expressed or implied, of the sponsors.

\bibliographystyle{unsrt}
\bibliography{references}

\end{document}

%% file: abstract.tex
Underwater robotic grasping is difficult due to degraded, highly variable imagery and the expense of collecting diverse underwater demonstrations. We introduce a system that (i) autonomously collects successful underwater grasp demonstrations via a self-supervised data collection pipeline and (ii) transfers grasp knowledge from on-land human demonstrations through a depth-based affordance representation that bridges the on-land–to–underwater domain gap and is robust to lighting and color shift. An affordance model trained on on-land handheld demonstrations is deployed underwater zero-shot via geometric alignment, and an affordance-conditioned diffusion policy is then trained on underwater demonstrations to generate control actions. In pool experiments, our approach improves grasping performance and robustness to background shifts, and enables generalization to objects seen only in on-land data, outperforming RGB-only baselines. 
Code, videos, and additional results are available at \url{https://umi-under-water.github.io}.

% \mc{notes for abstract\\}
% We present a solution that allows a subaquatic robot to locate and grasp objects using a combination of training from demonstration, from human demonstrations on land, and self-supervised grasping trials under water. The self-supervised trials use
% heuristic robot control and error recovery policies in combination with a light, mobile segmentation model to identify graspable objects.  

% After collecting 250 successful trials, a learned policy achieves $\approx 95$\% success rate for in domain.

% We use current and goal frames to learn an affordance heatmap to identify goal locations for grasping novel objects or selecting an object when multiple objects are in view. To incorporate data from human demonstrations, we introduce a cross-camera transformation to account for the very different optical conditions in air and under water. We fine-tune the resulting policy using the results of the self-supervised underwater model.
% \\
% Something about experiment setup, tests, results...